\documentclass[lettersize,journal]{IEEEtran}
\usepackage{amsmath,amsfonts}
\usepackage{algorithmic}
\usepackage{algorithm}
\usepackage{array}
\newcolumntype{L}[1]{>{\raggedright\arraybackslash}p{#1}}
\usepackage[caption=false,font=normalsize,labelfont=sf,textfont=sf]{subfig}
\usepackage{textcomp}
\usepackage{stfloats}
\usepackage{url}
\usepackage{verbatim}
\usepackage{graphicx}
\usepackage{cite}
\usepackage{booktabs}
\usepackage{multirow}
\usepackage[table]{xcolor}
\usepackage{todonotes}
\definecolor{oursblue}{RGB}{227,239,255}
\definecolor{groupgray}{RGB}{232,232,232}
\definecolor{baselinegray}{RGB}{242,242,242}
\definecolor{avggray}{RGB}{242,242,242}
\begin{document}
\title{PivotMerge: Bridging Heterogeneous Multimodal Pre-training via Post-Alignment Model Merging}

\author{
Zibo Shao\textsuperscript{*},
Baochen Xiong\textsuperscript{*},
Xiaoshan Yang,~\IEEEmembership{Member,~IEEE},
Yaguang Song,
Qimeng Zhang,
Haifeng Chen,
and Changsheng Xu,~\IEEEmembership{Fellow,~IEEE}
\thanks{\textsuperscript{*}Zibo Shao and Baochen Xiong contributed equally to this work.}
\thanks{\textit{Corresponding authors: Xiaoshan Yang and Yaguang Song.}}
\thanks{Zibo Shao, Baochen Xiong, Xiaoshan Yang, and Changsheng Xu are with the State Key Laboratory of Multimodal Artificial Intelligence Systems, Institute of Automation, Chinese Academy of Sciences, Beijing 100190, China, also with Pengcheng Laboratory, Shenzhen 518066, China, and also with the School of Artificial Intelligence, University of Chinese Academy of Sciences, Beijing 100049, China (e-mail: shaozibo2023@ia.ac.cn; xiongbaochen2022@ia.ac.cn; xiaoshan.yang@nlpr.ia.ac.cn; csxu@nlpr.ia.ac.cn).}
\thanks{Yaguang Song is with Pengcheng Laboratory, Shenzhen 518066, China (e-mail: songyg01@pcl.ac.cn).}
\thanks{Qimeng Zhang and Haifeng Chen are with the Data Science and Artificial Intelligence Research Institute, China United Network Communications Group Co., Ltd., Beijing 100033, China (e-mail: zhangqm50@chinaunicom.cn; chenhaifeng@chinaunicom.cn).}%
}


\maketitle

\begin{abstract}
Multimodal Large Language Models (MLLMs) rely on multimodal pre-training over diverse data sources, where different datasets often induce complementary cross-modal alignment capabilities. 
Model merging provides a cost-effective mechanism for integrating multiple expert MLLMs with complementary strengths into a unified model. However, existing model merging research mainly focuses on post-finetuning scenarios, leaving the pre-training stage largely unexplored.
We argue that the core of MLLM pre-training lies in establishing effective cross-modal alignment, which bridges visual and textual representations into a unified semantic space.
Motivated by this insight, we introduce the post-alignment merging task, which aims to integrate cross-modal alignment capabilities learned from heterogeneous multimodal pre-training. 
This setting introduces two key challenges: cross-domain parameter interference, where parameter updates learned from different data distributions conflict during merging, and layer-wise alignment contribution disparity, where different layers and projectors contribute unevenly to cross-modal alignment.
To address them, we propose \textbf{PivotMerge}, a post-alignment merging framework for cross-modal projectors. 
PivotMerge incorporates two key components: Shared-space Decomposition and Filtering, which disentangles shared alignment patterns from domain-specific variations and suppresses conflicting directions, and Alignment-guided Layer-wise Merging, which assigns layer-specific merging weights based on differing alignment contributions. 
We construct systematic CC12M-based post-alignment merging scenarios for evaluation. 
Extensive experiments on multiple multimodal benchmarks show that PivotMerge consistently outperforms existing baselines, demonstrating its effectiveness and generalization ability.
\end{abstract}


\section{Introduction}
\label{sec:Introduction}
\IEEEPARstart{I}{n} recent years, model merging~\cite{garipov2018loss, draxler2018essentially, wortsman2022model, choshen2022fusing} has emerged as an effective post-training paradigm for integrating multiple expert models with minimal computational cost and with little or no reliance on additional labeled training data~\cite{yang2023adamerging}. 
By directly combining model parameters or parameter-efficient components through training-free operations in weight space, model merging enables efficient reuse and aggregation of capabilities learned from different tasks, datasets, or training objectives, avoiding the substantial expense of retraining large-scale models~\cite{stoica2023zipit, ilharco2022editing, yadav2023ties, ortiz2023task}.

Recently, multimodal large language models (MLLMs)~\cite{liu2023visual, li2023blip, zhu2023minigpt, dai2023instructblip, li2024llava, wang2025internvl3, bai2025qwen3}, which extend LLMs with broader reasoning and perception capabilities through large-scale multimodal training, have gained substantial traction.
In this context, model merging provides a cost-effective mechanism to integrate multiple expert MLLMs with complementary strengths into a unified model without additional retraining.
Prior studies have explored this direction from several perspectives, including merging models trained on different modalities or multimodal tasks, composing existing MLLMs for capability enhancement or modality expansion~\cite{sung2023empirical, shukor2023unival, chen2024model}, and developing more adaptive or robust post-finetuning merging strategies~\cite{chen2024enhancing, wei2025unifying, du2025adamms, zeng2025robustmerge, qu2025uq, ma2026and}.
Despite this progress, existing multimodal model merging studies have predominantly focused on post-finetuning settings, while the multimodal pre-training stage remains largely unexplored from a model merging perspective.
\begin{figure}[t]
\centering
\includegraphics[width=\linewidth]{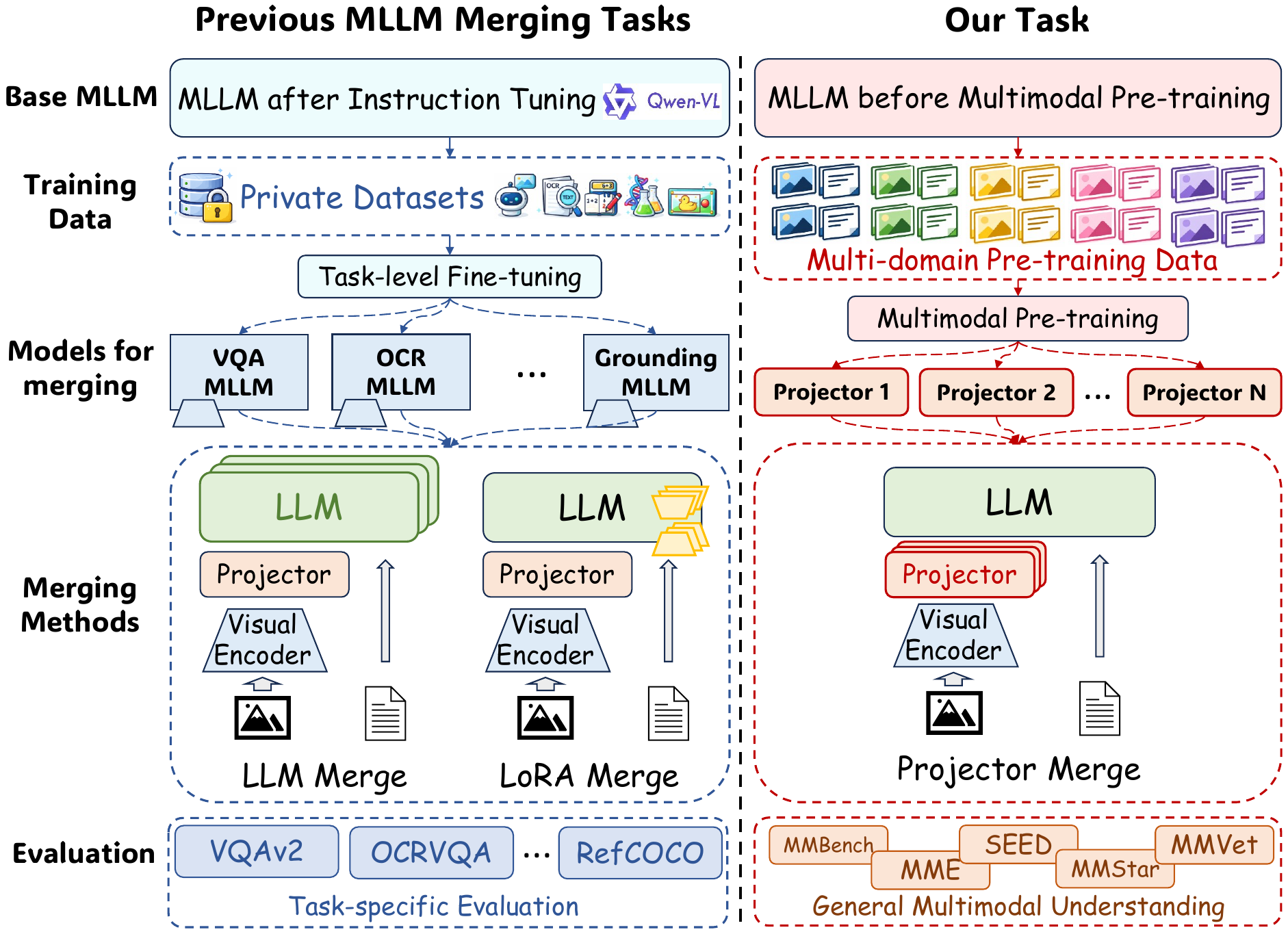}
\caption{Comparison between prior post-finetuning MLLM merging tasks and our post-alignment merging task.}
\label{fig:task_compare}
\vspace{-2ex}
\end{figure}

MLLMs obtained through multimodal pre-training on different datasets may develop distinct fundamental multimodal understanding capabilities. Performing model merging at this stage can help integrate these capabilities and provide a strong foundation for subsequent fine-tuning.
In this paper, we take an initial step toward model merging at the multimodal pre-training stage for MLLMs.
We argue that the core of MLLM pre-training lies in establishing effective cross-modal alignment, which bridges visual and textual representations into a unified semantic space for coherent multimodal understanding.
Motivated by this insight, we study model merging at the cross-modal alignment stage of multimodal pre-training.
Given the prohibitive cost of end-to-end pre-training for entire MLLMs, we adopt the widely studied “Visual-Projector-Language” architecture~\cite{liu2023visual, li2023blip, zhu2023minigpt, dai2023instructblip, chen2024expanding, bai2025qwen3} to instantiate the problem and conduct preliminary benchmarks.
We formally introduce the \textbf{post-alignment merging task} for bridging heterogeneous multimodal pre-training.
In this task, the visual encoder and LLM remain frozen, and only the lightweight cross-modal projectors, separately trained on heterogeneous image--text data distributions, are merged.
This setup enables effective integration of the cross-modal alignment capabilities learned by different projectors at the post-alignment stage, thereby providing a stronger foundation for subsequent fine-tuning.

Figure~\ref{fig:task_compare} highlights the key differences between prior post-finetuning MLLM merging and our post-alignment merging task, mainly in model acquisition and the scale of merged parameters.
In prior post-finetuning merging, models are typically obtained through further fine-tuning on different downstream tasks, and the merged components are usually backbone modules. 
In contrast, in our post-alignment merging setting, projectors are learned via multimodal pre-training on heterogeneous data distributions, and the merging targets are cross-modal projectors.
This task introduces several intrinsic challenges:
\textbf{(i) Cross-domain Parameter Interference.}
Parameter updates learned from different pre-training data distributions often conflict during merging.
Although their principal components tend to capture shared cross-modal alignment patterns, they also contain domain-specific variations induced by discrepancies in the training data, which introduce substantial interference into the merged parameters.
\textbf{(ii) Layer-wise Alignment Contribution Disparity.}
For a multi-layer projector, each layer contributes differently to cross-modal alignment. Moreover, at the same layer, different models can exhibit different cross-modal alignment capabilities. Therefore, projector merging should be adaptively performed in an alignment-aware manner across both layers and models.

To take a step toward addressing the proposed task, we propose \textbf{PivotMerge}, a post-alignment merging framework designed to bridge heterogeneous multimodal pre-training.
To address cross-domain parameter interference, PivotMerge employs \textbf{Shared-space Decomposition and Filtering}.
We first jointly decompose all projector updates into a shared orthogonal basis and projector-specific coefficients, mapping them into a shared coordinate system that mitigates parameter conflicts.
However, these coefficient matrices still entangle shared cross-modal alignment patterns with domain-specific variations. To suppress such interference, we propose to decouple each coefficient matrix into a primary alignment core and a domain-specific residual.
The former captures the dominant low-rank structure shared across projectors, reflecting common cross-modal alignment patterns, while the latter encodes unshared, idiosyncratic variations.
Since interference mainly arises from inconsistent residuals, we further introduce consistency-aware residual filtering to suppress directions with low cross-projector agreement. This design preserves the shared alignment structure while attenuating domain-specific conflicts, thereby more effectively mitigating cross-domain parameter interference.
To address layer-wise alignment contribution disparity, we introduce \textbf{Alignment-guided Layer-wise Merging}, which estimates the relative cross-modal alignment contributions of different layers across projectors and assigns corresponding layer-specific merging weights to the primary alignment cores. 
In this way, the differing alignment contributions of individual layers are explicitly reflected in the merging process.

Our main contributions can be summarized as follows:

\begin{itemize}
    \item We formulate a new post-alignment merging task for MLLMs, which enables effective integration of the complementary cross-modal alignment capabilities learned by different projectors at the post-alignment stage, thereby providing a stronger foundation for subsequent fine-tuning.
    
    \item We propose \textbf{PivotMerge}, a post-alignment merging framework for cross-modal projectors. Specifically, PivotMerge incorporates two key components: Shared-space Decomposition and Filtering, which mitigates cross-domain parameter interference, and Alignment-guided Layer-wise Merging, which adaptively assigns layer-specific merging weights based on differing alignment contributions.

    \item We construct systematic CC12M-based post-alignment merging scenarios for evaluation. Extensive experiments show that PivotMerge consistently outperforms existing baselines across multiple multimodal understanding benchmarks.
\end{itemize}

\section{Related Work}
\subsection{Model Merging}

Model merging has recently emerged as an efficient paradigm for integrating multiple expert models without additional training.
Instead of retraining a unified model from scratch, it combines the parameters or task-specific updates of independently fine-tuned models, enabling the merged model to inherit complementary abilities while maintaining the efficiency of a single model.
Early studies mainly rely on weight interpolation or averaging.
Representative examples include Model Soup~\cite{wortsman2022model}, which averages multiple fine-tuned checkpoints to improve robustness and generalization, and Task Arithmetic~\cite{ilharco2022editing}, which introduces task vectors as differences between fine-tuned and pre-trained weights, enabling capability composition via parameter-space arithmetic.
Subsequent works aim to reduce destructive interference among task updates.
For example, TIES-Merging~\cite{yadav2023ties} removes small-magnitude parameters and resolves sign conflicts, while DARE~\cite{yu2024language} adopts a drop-and-rescale strategy to stabilize merging.
Another line of work improves merging quality through parameter alignment or subspace transformation.
Git Re-Basin~\cite{ainsworth2022git} aligns neurons via permutation matching, while KnOTS~\cite{stoica2024model} performs SVD-based subspace alignment to improve compatibility.
More recent studies extend model merging to more challenging settings.
For instance, FuseLLM~\cite{wan2024knowledge} introduces token alignment and knowledge distillation to go beyond simple averaging, albeit with additional training cost, while EvoGM~\cite{jiang2026evogm} explores evolutionary generative optimization for adaptive merging coefficient search.
Adaptive composition has also been studied at the system level for LLM-based agents~\cite{xu2026evomas}.

In contrast to these works, which mainly target general model merging scenarios, our method is designed for multimodal merging, with a focus on cross-modal alignment.

\subsection{Model Merging in MLLMs}

With the rapid advancement of Multimodal Large Language Models (MLLMs) and related studies on multimodal pre-training and reasoning~\cite{zhuang2026cluster, zhuang2025grounding, tu2025perception}, model merging techniques have been extended to the multimodal domain to enable cross-modal knowledge transfer and unified representations. Early studies~\cite{sung2023empirical} validated the feasibility of merging Transformer models trained on different modalities, while UnIVAL~\cite{shukor2023unival} explored architectural unification across image, video, audio, and language tasks.
Recent works focus on enhancing perception or integrating modalities by composing off-the-shelf MLLMs. VisionFuse~\cite{chen2024enhancing} proposes a training-free framework that improves visual perception by combining multiple vision encoders, while DAMC~\cite{chen2024model} composes modality encoders and merges LLM parameters to mitigate interference. OptMerge~\cite{wei2025unifying} introduces a benchmark for MLLM merging and demonstrates unified multimodal capability via parameter merging.
More advanced strategies address architectural heterogeneity and merging reliability. AdaMMS~\cite{du2025adamms} handles heterogeneous architectures via unsupervised coefficient optimization, RobustMerge~\cite{zeng2025robustmerge} improves the robustness of merging parameter-efficient modules such as LoRA~\cite{hu2022lora}, and UQ-Merge~\cite{qu2025uq} introduces uncertainty-guided model selection and ordering.

Despite these advances, existing methods mainly focus on post-finetuning merging, largely overlooking the multimodal pre-training stage where fundamental cross-modal alignment is established. Our work instead studies post-alignment merging to integrate complementary alignment capabilities learned from heterogeneous multimodal pre-training.
\section{Methodology}
\label{sec:method}

\subsection{Task Definition}
\label{sec:method:problem}

In this section, we formalize the post-alignment merging task, including its setting, constraints, and evaluation principle.

We consider a Multimodal Large Language Model (MLLM) composed of three components: (i) a frozen vision encoder $f_v(\cdot)$, (ii) a frozen LLM $f_l(\cdot)$, and (iii) a trainable cross-modal projector $f_p(\cdot; \mathbf{W})$ parameterized by $\mathbf{W}$, which maps visual features into the language embedding space. Given an input image $\mathbf{X}^v$, the vision encoder extracts visual features $\mathbf{Z}^v = f_v(\mathbf{X}^v)$, which are then projected into visual tokens by
\begin{equation}
\mathbf{H}^v = f_p(\mathbf{Z}^v; \mathbf{W}) \in \mathbb{R}^{P \times C},
\end{equation}
where $P$ denotes the number of projected visual tokens and $C$ is the hidden size of the language model. The projected tokens $\mathbf{H}^v$ are then fed into the frozen LLM $f_l(\cdot)$ for response generation.

Let $\{\mathcal{D}_i\}_{i=1}^N$ denote $N$ large-scale image--text datasets with different distributions, where each sample is a pair $(\mathbf{X}^v, \mathbf{X}^t)$ and $\mathbf{X}^t = (\mathbf{x}_1^t,\ldots,\mathbf{x}_T^t)$ is the target text token sequence of length $T$ paired with $\mathbf{X}^v$.
The projector is trained by minimizing the standard autoregressive language modeling loss:
\begin{equation}
\mathcal{L}_{\mathcal{D}_i}(\mathbf{W})
=
\mathbb{E}_{(\mathbf{X}^v,\mathbf{X}^t)\sim \mathcal{D}_i}
\left[
-\sum_{j=1}^{T}
\log p\big(\mathbf{x}_j^t \mid \mathbf{x}_{<j}^t, \mathbf{H}^v\big)
\right],
\end{equation}
where $\mathbf{x}_j^t$ denotes the $j$-th text token, and $\mathbf{x}_{<j}^t$ denotes all preceding text tokens before position $j$.

Starting from a shared initialization $\mathbf{W}_0$, we independently optimize the projector on each dataset $\mathcal{D}_i$, yielding trained projectors $\{\mathbf{W}_1,\ldots,\mathbf{W}_N\}$.
Each $\mathbf{W}_i$ reflects the cross-modal alignment patterns learned from dataset $\mathcal{D}_i$, with parameter update $\Delta \mathbf{W}_i = \mathbf{W}_i - \mathbf{W}_0$.

The goal of \textbf{post-alignment merging} is to construct merged projector parameters $\mathbf{W}^\ast$ by merging $\{\mathbf{W}_1, \ldots, \mathbf{W}_N\}$ in parameter space, such that $\mathbf{W}^\ast$ integrates complementary alignment capabilities learned from different datasets.

We consider a training-free setting, where merging is performed directly in parameter space without gradient-based retraining.
The merged projector $\mathbf{W}^\ast$ is then plugged into the original MLLM and evaluated on multimodal understanding benchmarks without further fine-tuning, where downstream performance serves as a proxy for the quality and generalization of cross-modal alignment.

\subsection{Framework Overview}
\label{sec:method:overview}
\begin{figure*}[t]
\centering
\includegraphics[width=\textwidth]{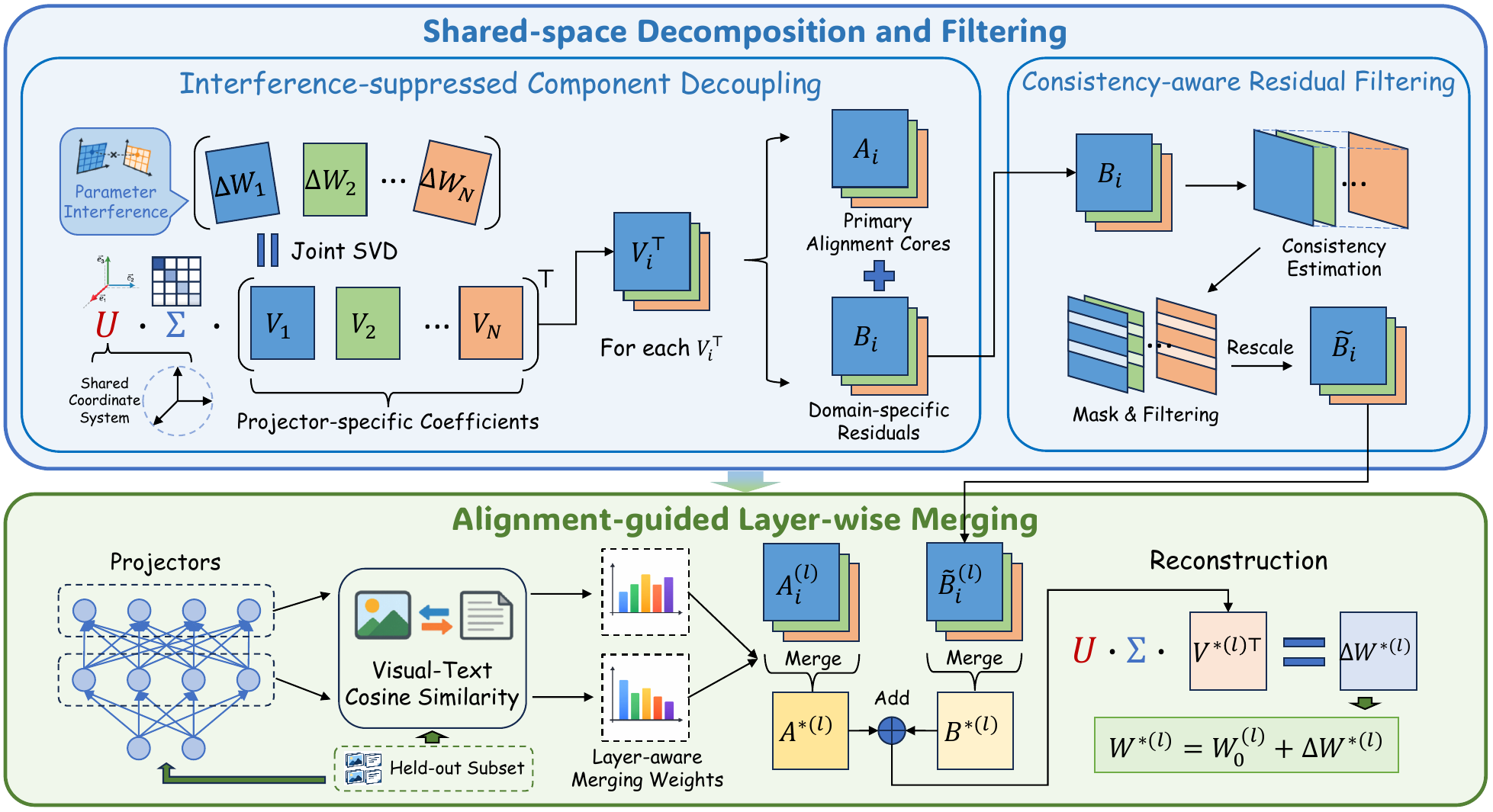}
\caption{
Overview of \textbf{PivotMerge}.
PivotMerge introduces two modules to address the key challenges of post-alignment merging: Shared-space Decomposition and Filtering mitigates cross-domain parameter interference, and Alignment-guided Layer-wise Merging addresses layer-wise alignment contribution disparity.
}
\label{fig:method_overview}
\vspace{-2ex}
\end{figure*}

We propose a post-alignment merging framework, termed PivotMerge, for cross-modal projectors learned during multimodal pre-training, as illustrated in Fig.~\ref{fig:method_overview}.  
PivotMerge consists of two modules:
\textbf{Shared-space Decomposition and Filtering}, which performs joint decomposition and component-wise decoupling with conflict-aware residual filtering to mitigate cross-domain parameter interference; and
\textbf{Alignment-guided Layer-wise Merging}, which assigns layer-specific merging weights to address layer-wise alignment contribution disparity.

\subsection{Shared-space Decomposition and Filtering}
\label{sec:method:saf}

This module mitigates cross-domain parameter interference through two stages: \textbf{Interference-suppressed Component Decoupling} and \textbf{Consistency-aware Residual Filtering}.
For clarity, we omit the layer index $\ell$ in the following derivation. The same decomposition, filtering, and merging procedure is applied independently to each projector layer.

\subsubsection{Interference-suppressed Component Decoupling}
\label{sec:method:decouple}

Directly merging projectors in the original parameter space can be unstable, as heterogeneous pre-training data distributions induce substantial interference among updates.
To address this, we express all projector updates in a shared coordinate system and decouple them into shared and domain-specific components.

Following the definition in Section~\ref{sec:method:problem}, $\{\Delta \mathbf{W}_1, \Delta \mathbf{W}_2, \ldots, \Delta \mathbf{W}_N\}$ denote the projector updates. 
We first perform \textit{Joint Decomposition} by concatenating these updates and applying joint singular value decomposition following~\cite{stoica2024model}:
\begin{equation}
\begin{aligned}
    \big[ \Delta \mathbf{W}_1, \Delta \mathbf{W}_2, \ldots, \Delta \mathbf{W}_N \big]
    &= \mathbf{U} \mathbf{\Sigma} \mathbf{V}^\top \\
    &= \mathbf{U} \mathbf{\Sigma} \big[ \mathbf{V}_1^\top, \mathbf{V}_2^\top, \ldots, \mathbf{V}_N^\top \big],
\end{aligned}
\end{equation}
where $\mathbf{U}$ is a shared orthogonal basis, $\mathbf{\Sigma}$ contains singular values, and $\mathbf{V}^\top$ is partitioned into projector-specific coefficient matrices
$\{\mathbf{V}_1^\top, \ldots, \mathbf{V}_N^\top\}$.
Each projector update can then be written as
\begin{equation}
    \Delta \mathbf{W}_i = \mathbf{U} \mathbf{\Sigma} \mathbf{V}_i^\top.
\end{equation}

This formulation places heterogeneous updates into a shared space, where $\mathbf{V}_i^\top$ characterizes each projector under a shared basis.
However, these coefficient matrices still mix shared alignment patterns with domain-specific variations induced by discrepancies in the training data.
We therefore further perform \textit{Component-wise Decoupling} on each coefficient matrix into a primary alignment core and a domain-specific residual.

Specifically, for each $\mathbf{V}_i^\top$, we perform a second-stage singular value decomposition:
\begin{equation}
\mathbf{V}_i^\top = \mathbf{Q}_i \boldsymbol{\Lambda}_i \mathbf{R}_i^\top,
\end{equation}
and retain the top-$r$ components to construct the primary alignment core:
\begin{equation}
\mathbf{A}_i = \mathbf{Q}_i \boldsymbol{\Lambda}_i^{(r)} \mathbf{R}_i^\top,
\end{equation}
where $\boldsymbol{\Lambda}_i^{(r)}$ keeps the largest $r$ singular values.
The remaining part is defined as the domain-specific residual:
\begin{equation}
\mathbf{B}_i = \mathbf{V}_i^\top - \mathbf{A}_i.
\end{equation}

Here, $\mathbf{A}_i$ captures dominant low-rank structures shared across projectors, reflecting common cross-modal alignment patterns, while $\mathbf{B}_i$ retains residual variations from domain-specific data preferences.
This decoupling reduces interference among updates and yields cleaner residuals for subsequent filtering.

\subsubsection{Consistency-aware Residual Filtering}
\label{sec:method:residual_filtering}

We assume that the main conflicts arise from inconsistencies in the domain-specific residuals, which capture unshared data preferences.
To mitigate this, we apply consistency-aware filtering only to the residuals $\mathbf{B}_i$, while preserving the primary alignment cores $\mathbf{A}_i$.

For the $k$-th row of $\mathbf{B}_i$, we collect
$\{\mathbf{b}_{1,k}, \mathbf{b}_{2,k}, \ldots, \mathbf{b}_{N,k}\}$ and compute their average pairwise cosine similarity:
\begin{equation}
c_k
=
\frac{1}{N(N-1)}
\sum_{\substack{i,j=1 \\ i \neq j}}^{N}
\cos(\mathbf{b}_{i,k}, \mathbf{b}_{j,k}).
\end{equation}

Based on $c_k$, we construct a mask vector $\mathbf{m}$ with $m_k = \mathrm{sigmoid}\big(\gamma (c_k - \tau)\big)$, where $\gamma$ controls the sharpness of the sigmoid function and $\tau$ is the consistency threshold.
We then apply row-wise masking to the residual:
\begin{equation}
\mathbf{B}_i^{\text{masked}} = \mathbf{m} \odot \mathbf{B}_i,
\end{equation}
where $\odot$ denotes row-wise multiplication between the mask vector and the residual matrix.
In this way, residual directions with low cross-projector consistency receive smaller mask values and are correspondingly suppressed.
As a result, inconsistent and potentially conflicting components in the domain-specific residuals are attenuated before merging.

To preserve the overall magnitude, we apply $L_1$-norm compensation:
\begin{equation}
\widetilde{\mathbf{B}}_i
=
\mathbf{B}_i^{\text{masked}}
\cdot
\frac{\|\mathbf{B}_i\|_1}{\|\mathbf{B}_i^{\text{masked}}\|_1}.
\end{equation}

This rescaling preserves the residual magnitude while maintaining the suppression effect.
The primary alignment cores $\mathbf{A}_i$ and filtered residuals $\widetilde{\mathbf{B}}_i$ are then used in the subsequent Alignment-guided Layer-wise Merging stage.

\subsection{Alignment-guided Layer-wise Merging}
\label{sec:method:layeraware}

Projectors exhibit layer-wise alignment contribution disparity, since different layers contribute unevenly to cross-modal alignment, and even at the same layer, different projectors may exhibit different alignment capabilities.
To address this disparity, we introduce Alignment-guided Layer-wise Merging, which assigns layer-specific merging weights to the primary alignment cores based on their differing alignment contributions.

For an $L$-layer projector, we denote its parameters by
$\mathbf{W}_i=\{\mathbf{W}_i^{(\ell)}\}_{\ell=1}^L$,
where $\mathbf{W}_i^{(\ell)}$ is the weight matrix of layer $\ell \in \{1,\ldots,L\}$ for projector $i \in \{1,\ldots,N\}$.
After Shared-space Decomposition and Filtering, each projector at layer $\ell$ is represented by a primary alignment core $\mathbf{A}_i^{(\ell)}$ and a filtered domain-specific residual $\widetilde{\mathbf{B}}_i^{(\ell)}$.

\subsubsection{Layer-wise Alignment Weighting of Primary Alignment Cores}
To reflect the relative cross-modal alignment capability of different layers across projectors, we estimate a layer-specific cross-modal alignment score for each projector on its own held-out validation subset $\mathcal{D}_i^{\mathrm{val}}$, retained from the end of the pre-training data of projector $\mathbf{W}_i$:
\begin{equation}
s_i^{(\ell)}
=
\frac{1}{|\mathcal{D}_i^{\mathrm{val}}|}
\sum_{(\mathbf{X}^v, \mathbf{X}^t)\in\mathcal{D}_i^{\mathrm{val}}}
\cos\!\Big(
f_p^{(\ell)}\big(f_v(\mathbf{X}^v); \mathbf{W}_i\big), \,
\mathbf{e}(\mathbf{X}^t)
\Big),
\end{equation}
where $f_p^{(\ell)}(\cdot; \mathbf{W}_i)$ denotes the intermediate projected features extracted from layer $\ell$ of projector $\mathbf{W}_i$, and $\mathbf{e}(\mathbf{X}^t)$ denotes the text embedding of the target text sequence $\mathbf{X}^t$.

To reflect the effective contribution of the current layer, we use the layer-wise score increment rather than the absolute score. For $\ell > 1$, we define
\begin{equation}
\Delta s_i^{(\ell)} = s_i^{(\ell)} - s_i^{(\ell-1)},
\end{equation}
and for the first layer use $\Delta s_i^{(1)} = s_i^{(1)}$.
The layer-specific merging weight for projector $i$ at layer $\ell$ is then computed as
\begin{equation}
\alpha_i^{(\ell)}
=
\frac{\exp\!\left(\Delta s_i^{(\ell)}/\beta\right)}
{\sum_{j=1}^{N}\exp\!\left(\Delta s_j^{(\ell)}/\beta\right)},
\end{equation}
where $\beta$ is the softmax temperature controlling the sharpness of the weight distribution.

These weights are used only for merging the primary alignment cores:
\begin{equation}
\mathbf{A}^{*(\ell)}
=
\mathrm{Merge}\!\left(
\left\{\mathbf{A}_i^{(\ell)}\right\}_{i=1}^{N},
\left\{\alpha_i^{(\ell)}\right\}_{i=1}^{N}
\right),
\end{equation}
where $\mathrm{Merge}(\cdot)$ denotes a merging operator instantiated by an existing method (e.g., TIES).

\begin{algorithm}[t]
\caption{PivotMerge}
\label{alg:our_method}
\begin{algorithmic}[1]
\REQUIRE Projectors $\{\mathbf{W}_i=\{\mathbf{W}_i^{(\ell)}\}_{\ell=1}^L\}_{i=1}^N$, shared initialization $\mathbf{W}_0=\{\mathbf{W}_0^{(\ell)}\}_{\ell=1}^L$, pre-computed layer-wise score increments $\{\Delta s_i^{(\ell)}\}$, number of layers $L$, rank $r$, sharpness $\gamma$, threshold $\tau$, temperature $\beta$
\ENSURE Full merged projector $\mathbf{W}^\ast=\{\mathbf{W}^{*(\ell)}\}_{\ell=1}^L$

\FOR{each layer $\ell = 1,\dots,L$}
    \STATE \textcolor{gray}{\emph{Shared-space Decomposition and Filtering}}
    \STATE $\Delta \mathbf{W}_i^{(\ell)} \leftarrow \mathbf{W}_i^{(\ell)} - \mathbf{W}_0^{(\ell)}$, for $i=1,\dots,N$
    \STATE Perform joint SVD on $[\Delta \mathbf{W}_1^{(\ell)}, \dots, \Delta \mathbf{W}_N^{(\ell)}]$
    \STATE Obtain $\mathbf{U}^{(\ell)}$, $\boldsymbol{\Sigma}^{(\ell)}$, and $\{\mathbf{V}_i^{(\ell)\top}\}_{i=1}^N$

    \FOR{each projector $i = 1,\dots,N$}
        \STATE Perform SVD on $\mathbf{V}_i^{(\ell)\top}$:
        $\mathbf{V}_i^{(\ell)\top}=\mathbf{Q}_i^{(\ell)}\boldsymbol{\Lambda}_i^{(\ell)}\mathbf{R}_i^{(\ell)\top}$
        \STATE $\mathbf{A}_i^{(\ell)} \leftarrow \mathbf{Q}_i^{(\ell)}\big[\boldsymbol{\Lambda}_i^{(\ell)}\big]^{(r)}\mathbf{R}_i^{(\ell)\top}$
        \STATE $\mathbf{B}_i^{(\ell)} \leftarrow \mathbf{V}_i^{(\ell)\top} - \mathbf{A}_i^{(\ell)}$
    \ENDFOR
    \STATE $c_k^{(\ell)} \leftarrow
    \frac{1}{N(N-1)}
    \sum_{i \neq j}
    \cos(\mathbf{b}_{i,k}^{(\ell)}, \mathbf{b}_{j,k}^{(\ell)})$,
    for each row $k$
    \STATE Compute mask entries $m_k^{(\ell)} = \mathrm{sigmoid}\!\big(\gamma(c_k^{(\ell)}-\tau)\big)$

    \FOR{each projector $i = 1,\dots,N$}
        \STATE $\mathbf{B}_i^{(\ell),\text{masked}} \leftarrow \mathbf{m}^{(\ell)} \odot \mathbf{B}_i^{(\ell)}$
        \STATE $\widetilde{\mathbf{B}}_i^{(\ell)} \leftarrow \mathbf{B}_i^{(\ell),\text{masked}}
        \cdot
        \frac{\|\mathbf{B}_i^{(\ell)}\|_1}{\|\mathbf{B}_i^{(\ell),\text{masked}}\|_1}$
    \ENDFOR

    \STATE \textcolor{gray}{\emph{Alignment-guided Layer-wise Merging}}
    \STATE $\alpha_i^{(\ell)} \leftarrow
    \frac{\exp\!\left(\Delta s_i^{(\ell)}/\beta\right)}
    {\sum_{j=1}^{N}\exp\!\left(\Delta s_j^{(\ell)}/\beta\right)}$,
    for $i=1,\dots,N$
    \STATE $\mathbf{A}^{*(\ell)} \leftarrow \mathrm{Merge}\!\left(\{\mathbf{A}_i^{(\ell)}\}_{i=1}^N,\{\alpha_i^{(\ell)}\}_{i=1}^N\right)$
    \STATE $\mathbf{B}^{*(\ell)} \leftarrow \mathrm{Merge}\!\left(\{\widetilde{\mathbf{B}}_i^{(\ell)}\}_{i=1}^N\right)$
    \STATE $\mathbf{V}^{*(\ell)\top} \leftarrow \mathbf{A}^{*(\ell)} + \mathbf{B}^{*(\ell)}$
    \STATE $\mathbf{W}^{*(\ell)} \leftarrow \mathbf{W}_0^{(\ell)} + \mathbf{U}^{(\ell)}\boldsymbol{\Sigma}^{(\ell)}\mathbf{V}^{*(\ell)\top}$
\ENDFOR

\RETURN $\mathbf{W}^\ast=\{\mathbf{W}^{*(\ell)}\}_{\ell=1}^L$
\end{algorithmic}
\end{algorithm}

\subsubsection{Residual Merging and Reconstruction}
The filtered domain-specific residuals are merged using a unified set of weights shared across projectors, without applying the cross-modal alignment scores:
\begin{equation}
\mathbf{B}^{*(\ell)}
=
\mathrm{Merge}\!\left(
\left\{\widetilde{\mathbf{B}}_i^{(\ell)}\right\}_{i=1}^{N}
\right).
\end{equation}

The final merged coefficient matrix at layer $\ell$ is $\mathbf{V}^{*(\ell)\top} = \mathbf{A}^{*(\ell)} + \mathbf{B}^{*(\ell)}$.
The merged parameter is then reconstructed as
\begin{equation}
\mathbf{W}^{*(\ell)}
=
\mathbf{W}_0^{(\ell)}
+
\mathbf{U}^{(\ell)}
\boldsymbol{\Sigma}^{(\ell)}
\mathbf{V}^{*(\ell)\top},
\end{equation}
where $\mathbf{W}_0^{(\ell)}$ is the initialization parameter of layer $\ell$, and $\mathbf{U}^{(\ell)}$ and $\boldsymbol{\Sigma}^{(\ell)}$ are the shared orthogonal basis and singular value matrix obtained from Joint Decomposition.
The full merged projector is therefore given by
$\mathbf{W}^\ast = \{\mathbf{W}^{*(\ell)}\}_{\ell=1}^L$.

The overall procedure of PivotMerge is summarized in Algorithm~\ref{alg:our_method}.

\section{Experiments}
\label{sec:experiments}

\subsection{Experimental Setup}

\subsubsection{Models}
\label{sec:exp:models}

We conduct all experiments based on LLaVA-1.5-7B~\cite{liu2024improved}.
Specifically, the visual encoder is CLIP-ViT-L/14-336px~\cite{radford2021learning}, and the language model is Vicuna-7B~\cite{vicuna2023}.
For cross-modal projection, we adopt a two-layer MLP projector with GELU activation~\cite{liu2024improved}.
Following the pre-training protocol of LLaVA~\cite{liu2023visual} and the setting defined in Section~\ref{sec:method:problem}, we keep both the visual encoder and the LLM frozen, and train only the cross-modal projector.
See the \textit{Implementation Details} section in the supplementary material for more details.

\subsubsection{Dataset}
\label{sec:exp:dataset}

Inspired by the cross-modal pre-training strategy in LLaVA, we construct two groups of multimodal pre-training splits based on CC12M~\cite{changpinyo2021conceptual}, where each split is a subset of CC12M. Each group consists of five disjoint splits, each containing 558K image--text pairs.

To investigate the effect of data distribution discrepancy on post-alignment merging, we construct two experimental settings with different partition strategies: \textbf{Clustered Setting} and \textbf{Random Setting}. 
In the Clustered Setting, the CC12M corpus is partitioned into five splits via feature-based clustering according to semantic similarity in the joint visual--textual space, leading to larger distributional divergence across splits. 
In the Random Setting, the CC12M corpus is partitioned into five equally sized splits by random sampling, resulting in comparatively milder domain shifts.

\begin{figure}[t]
\centering
\includegraphics[width=\linewidth]{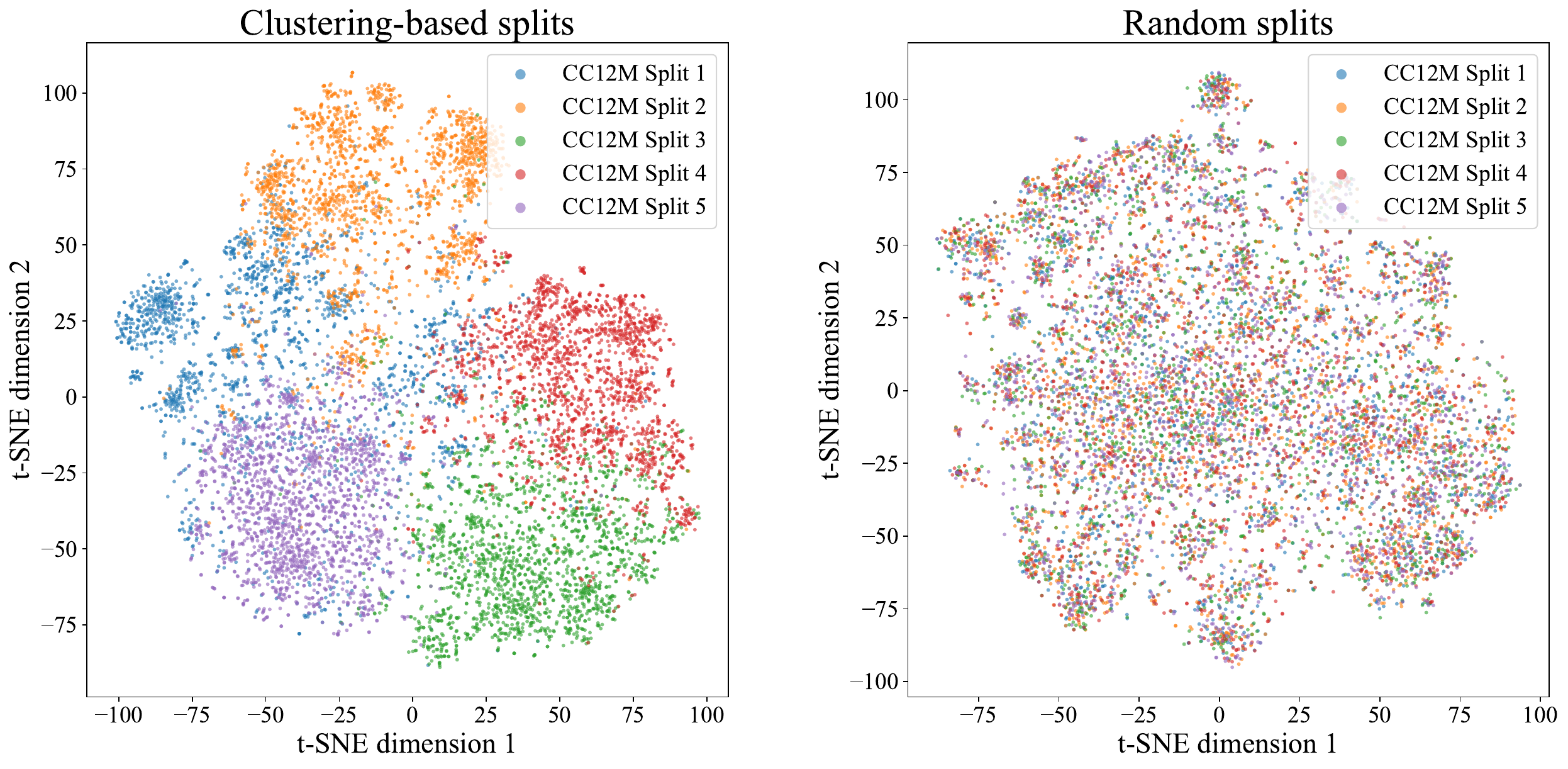}
\caption{
Visualization of the visual feature distributions of the five CC12M splits using CLIP embeddings and t-SNE. 
The left panel shows the clustered splits, while the right panel shows the random splits. 
Each point represents an image sample, and colors denote different splits. 
The clustered partition exhibits clearly separated distributions, whereas the random partition leads to largely overlapping splits, indicating smaller distribution discrepancies.
}
\label{fig:cc12m_splits_tsne}
\vspace{-2ex}
\end{figure}

These two strategies yield pre-training splits with different levels of inter-split heterogeneity. Each split is used to independently pre-train the projector of LLaVA, producing a set of projectors for subsequent merging. To illustrate the distributional differences, we visualize the visual feature distributions using CLIP~\cite{radford2021learning} embeddings followed by t-SNE~\cite{van2008visualizing}, as shown in Fig.~\ref{fig:cc12m_splits_tsne}.

\begin{table*}[t]
\centering
\caption{Performance comparison on multimodal benchmarks. 
The single models are trained on five splits of CC12M partitioned via \textbf{clustering}. 
Best results among merging methods are shown in bold and second-best results are underlined.}
\label{tab:main_results_cluster}
\resizebox{\textwidth}{!}{
\begin{tabular}{lccccccccc}
\toprule
Method 
& MMVet 
& MMBench$_{\text{EN}}$ 
& SEEDBench 
& LLaVABench 
& POPE 
& MME-P 
& MMVP 
& MMStar 
& \textbf{Avg.} \\
\midrule

\multicolumn{10}{c}{\textit{Single Models (Clustered Splits)}} \\
\midrule
CC12M Split 1 
& 26.3 & 15.2 & 11.9 & 33.1 & 41.8 & 666.8 & 15.3 & 20.0 & 24.6 \\
CC12M Split 2 
& 20.0 & 16.4 & 22.1 & 28.2 & 75.6 & 1037.4 & 23.3 & 25.7 & 32.9 \\
CC12M Split 3 
& 28.4 & 25.7 & 25.6 & 41.3 & 70.8 & 843.8 & 31.0 & 26.5 & 36.4 \\
CC12M Split 4 
& 17.5 & 17.0 & 22.5 & 33.0 & 64.9 & 585.0 & 23.0 & 22.9 & 28.8 \\
CC12M Split 5 
& 20.6 & 24.2 & 30.6 & 37.4 & 26.2 & 476.9 & 40.3 & 26.7 & 28.7 \\

\midrule
\multicolumn{10}{c}{\textit{Merging Methods}} \\
\midrule

\rowcolor{baselinegray}
Weight Average 
& 26.9 & 20.4 & 27.1 & 41.0 & 76.0 & 952.0 & 33.7 & 23.9 & 37.1 \\

\rowcolor{baselinegray}
Task Arithmetic 
& 21.6 & 18.3 & 27.5 & 37.1 & 68.5 & 702.5 & \underline{34.3} & 25.7 & 33.5 \\

\rowcolor{baselinegray}
TIES Merging 
& \underline{27.3} & \underline{31.2} & 28.9 & 40.5 & \textbf{77.3} & \underline{1073.4} & 32.7 & 25.1 & \underline{39.6} \\

\rowcolor{baselinegray}
TIES w/ DARE 
& 24.3 & 19.4 & 19.5 & \underline{41.5} & 68.6 & 555.2 & 20.7 & 23.7 & 30.7 \\

\rowcolor{baselinegray}
MetaGPT 
& 26.6 & 22.3 & 27.8 & 41.0 & \underline{76.5} & 966.4 & \textbf{34.7} & 23.3 & 37.6 \\

\rowcolor{baselinegray}
Iso-C 
& 8.2 & 2.3 & 8.9 & 17.6 & 27.8 & 141.0 & 28.3 & 13.5 & 14.2 \\

\rowcolor{baselinegray}
TSV-M 
& 23.2 & 24.2 & \underline{32.0} & 36.9 & 69.0 & 766.8 & 28.7 & \underline{27.3} & 35.0 \\

\midrule
\rowcolor{oursblue}
\textbf{PivotMerge} 
& \textbf{27.8} 
& \textbf{32.0} 
& \textbf{33.7} 
& \textbf{48.0} 
& 75.8 
& \textbf{1111.1} 
& \underline{34.3} 
& \textbf{27.5} 
& \textbf{41.8} \\
\bottomrule
\end{tabular}
}
\vspace{4pt}

\caption{Performance comparison on multimodal benchmarks. 
The single models are trained on five splits of CC12M partitioned via \textbf{random sampling}. 
Best results among merging methods are shown in bold and second-best results are underlined.}
\label{tab:main_results_random}
\resizebox{\textwidth}{!}{
\begin{tabular}{lccccccccc}
\toprule
Method 
& MMVet 
& MMBench$_{\text{EN}}$ 
& SEEDBench 
& LLaVABench 
& POPE 
& MME-P 
& MMVP 
& MMStar 
& \textbf{Avg.} \\
\midrule
\multicolumn{10}{c}{\textit{Single Models (Random Splits)}} \\
\midrule
CC12M Split 1 
& 26.9 & 28.3 & 24.6 & 39.3 & 77.6 & 1207.2 & 27.7 & 26.1 & 38.9 \\
CC12M Split 2 
& 24.2 & 22.8 & 16.9 & 35.8 & 72.9 & 891.3 & 35.3 & 28.0 & 35.1 \\
CC12M Split 3 
& 23.9 & 34.6 & 22.0 & 42.5 & 75.4 & 975.7 & 28.3 & 27.3 & 37.8 \\
CC12M Split 4 
& 25.0 & 35.8 & 22.3 & 38.1 & 76.0 & 1022.9 & 30.7 & 27.2 & 38.3 \\
CC12M Split 5 
& 23.1 & 34.0 & 17.4 & 35.2 & 72.1 & 831.1 & 34.7 & 27.8 & 35.7 \\
\midrule
\multicolumn{10}{c}{\textit{Merging Methods}} \\
\midrule

\rowcolor{baselinegray}
Weight Average 
& 25.0 & \underline{35.0} & 20.6 & \underline{40.2} & \textbf{77.1} & \underline{1078.2} & 35.0 & \underline{28.3} & \underline{39.4} \\

\rowcolor{baselinegray}
Task Arithmetic 
& 27.0 & 29.3 & \underline{29.1} & 38.4 & 68.5 & 709.7 & \textbf{40.0} & 26.0 & 36.7 \\

\rowcolor{baselinegray}
TIES Merging 
& \underline{28.7} & 32.3 & \textbf{31.2} & 39.2 & 68.9 & 754.4 & \textbf{40.0} & 27.9 & 38.2 \\

\rowcolor{baselinegray}
TIES w/ DARE 
& 17.7 & 21.0 & 3.3 & 26.5 & 22.2 & 137.7 & \underline{38.3} & 25.9 & 20.2 \\

\rowcolor{baselinegray}
MetaGPT 
& 25.3 & \textbf{35.6} & 20.6 & 39.5 & \underline{76.9} & 1072.5 & 35.0 & \textbf{28.5} & \underline{39.4} \\

\rowcolor{baselinegray}
Iso-C 
& 6.4 & 2.6 & 9.8 & 20.0 & 30.3 & 135.4 & 31.3 & 13.2 & 15.0 \\

\rowcolor{baselinegray}
TSV-M
& 27.8 & 28.6 & 27.0 & 40.1 & 67.5 & 740.4 & 21.3 & 17.9 & 33.4 \\

\midrule
\rowcolor{oursblue}
\textbf{PivotMerge} 
& \textbf{30.1} 
& 34.3 
& 21.7 
& \textbf{42.9} 
& 76.5 
& \textbf{1101.2} 
& 35.0 
& 27.3 
& \textbf{40.4} \\
\bottomrule
\end{tabular}
}
\vspace{-2ex}
\end{table*}

\subsubsection{Baselines}
\label{sec:exp:baselines}

We compare against representative model merging baselines commonly used in prior work, including Weight Averaging~\cite{wortsman2022model}, Task Arithmetic~\cite{ilharco2022editing}, TIES Merging~\cite{yadav2023ties}, DARE~\cite{yu2024language}, MetaGPT~\cite{wei2025unifying}, Iso-C~\cite{marczak2025no}, and TSV-M~\cite{gargiulo2025task}.

\subsubsection{Benchmarks}
\label{sec:exp:benchmarks}

We evaluate the merged models on a diverse suite of multimodal benchmarks, including MMVet~\cite{yu2023mm}, MMBench$_{\text{EN}}$~\cite{liu2024mmbench}, SEED~\cite{li2023seed}, LLaVABench~\cite{liu2023visual}, POPE~\cite{li2023evaluating}, MME~\cite{fu2023mme}, MMVP~\cite{tong2024eyes}, and MMStar~\cite{chen2024we}.
These benchmarks cover a broad range of multimodal understanding capabilities, providing a comprehensive evaluation of cross-modal alignment quality.
To ensure a comparable overall score, we normalize the MME-P results by dividing them by 20 when computing the average, as the total score of the MME perception benchmark is 2000.

\subsubsection{Merging Details}
\label{sec:exp:merging_details}

For Shared-space Decomposition and Filtering, we set the rank of the primary alignment core to $r=64$ in the component-wise decoupling stage, i.e., we retain the top 64 singular components of each coefficient matrix to form $\mathbf{A}_i$.
In the subsequent Consistency-aware Residual Filtering stage, we set $\gamma = 20.0$.

For Alignment-guided Layer-wise Merging, we reserve 1,000 held-out samples from the end of projector pre-training to estimate the cross-modal alignment scores for layer-wise weighting.
These scores determine the merging weights for the primary alignment cores $\mathbf{A}_i$, while the residual branch uses uniform averaging weights.
Unless otherwise specified, TIES~\cite{yadav2023ties} is adopted as the underlying merging operator in both branches.

\subsection{Overall Performance}

Table~\ref{tab:main_results_cluster} and Table~\ref{tab:main_results_random} present the overall results under the clustered and random settings. Across both settings, PivotMerge consistently achieves the best or highly competitive performance, validating the effectiveness of merging cross-modal projectors.

\textbf{Clustered Setting.}
Due to larger distribution discrepancies among splits, single models exhibit wider performance gaps, implying stronger conflicts among projectors learned from heterogeneous data. In this setting, the conflict-mitigation designs of PivotMerge become particularly effective. PivotMerge achieves the best average score of 41.8, surpassing TIES, the strongest baseline, by 2.2 points. It also attains the best results on six benchmarks and the second-best result on one benchmark, demonstrating its stronger ability to alleviate cross-model conflicts and integrate complementary cross-modal alignment capabilities.

\textbf{Random Setting.}
Performance differences among single models are smaller, indicating reduced diversity across expert models and thus less severe conflicts during merging. As a result, the room for improvement is more limited in this more homogeneous setting. Nevertheless, PivotMerge still achieves the best overall average (40.4) and the best performance on three benchmarks, showing that it remains effective even when domain discrepancies and cross-model conflicts are relatively mild.

\begin{table}[t]
\centering
\caption{Ablation study of each component.}
\label{tab:ablation}
\renewcommand{\arraystretch}{1.1}
\setlength{\tabcolsep}{3.7pt}
\begin{tabular}{ccccc|cccc|c}
\toprule
JD & CAS & CD & CRF & LM
& SEED & LLaVA & MMVP & MMStar
& \textbf{Avg} \\
\midrule

$\times$ & $\times$ & $\times$ & $\times$ & $\times$
& 28.9 & 40.5 & 32.7 & 25.1 & 31.8 \\

$\checkmark$ & $\times$ & $\times$ & $\times$ & $\times$
& 32.7 & 44.5 & 32.7 & 26.6 & 34.1 \\

$\times$ & $\checkmark$ & $\times$ & $\times$ & $\times$
& 29.2 & 42.9 & 31.7 & 24.8 & 32.2 \\

$\checkmark$ & $\checkmark$ & $\times$ & $\times$ & $\times$
& 35.1 & 44.6 & 34.7 & 26.8 & 35.3 \\

$\checkmark$ & $\checkmark$ & $\checkmark$ & $\times$ & $\times$
& 34.0 & 47.6 & 34.0 & 26.5 & 35.5 \\

$\checkmark$ & $\checkmark$ & $\checkmark$ & $\checkmark$ & $\times$
& 34.2 & 46.2 & 36.0 & 26.8 & 35.8 \\

$\checkmark$ & $\checkmark$ & $\checkmark$ & $\times$ & $\checkmark$
& 33.6 & 47.6 & 33.3 & 27.7 & 35.6 \\

$\checkmark$ & $\checkmark$ & $\checkmark$ & $\checkmark$ & $\checkmark$
& 33.7 & 48.0 & 34.3 & 27.5 & \textbf{35.9} \\

\bottomrule
\end{tabular}
\vspace{-3ex}
\end{table}

\subsection{Ablation Study}

We conduct ablation experiments under the clustered setting to evaluate the effectiveness of each component in PivotMerge, including Joint Decomposition (JD), cross-modal alignment score (CAS), Component-wise Decoupling (CD), Consistency-aware Residual Filtering (CRF), and Alignment-guided Layer-wise Merging (LM). 
When CAS is disabled, we replace it with uniform averaging weights across projectors. 
When CAS is enabled but CD is not applied, the alignment scores are directly used to weight the coefficient matrices $\mathbf{V}_i$ without decoupling. 
When both CAS and CD are used but LM is removed, we apply a shared merging weight across all layers, computed from the alignment score of the final projection layer. 
When CRF is disabled, residual components are directly merged without filtering. 
This design allows us to isolate the role of each component in a controlled manner.

Results are summarized in Table~\ref{tab:ablation}.
Without JD, the performance is consistently lower, indicating that directly merging parameters in the original space is insufficient to handle cross-domain parameter interference. 
Introducing JD leads to clear improvements across all benchmarks, demonstrating the importance of constructing a shared space for mitigating interference among heterogeneous updates.
Applying CAS further improves performance, suggesting that estimating cross-modal alignment capability provides a more informative merging signal than uniform averaging. 
However, without JD, the improvement brought by CAS remains limited, indicating that alignment-aware weighting is more effective when performed in the shared space constructed by JD.
With CD enabled, performance becomes more stable and improves across multiple benchmarks, showing that separating the primary alignment core from domain-specific residuals helps preserve shared cross-modal alignment patterns while reducing interference. 
Introducing CRF consistently improves performance, indicating that filtering inconsistent residual directions further stabilizes the merging process by suppressing conflicting domain-specific variations.
LM brings additional gains, particularly on benchmarks requiring complex reasoning, highlighting the importance of modeling layer-wise alignment contribution disparity. 
This suggests that treating all layers equally is suboptimal, and layer-aware weighting better reflects the functional roles of different projection layers.

\subsection{Analysis}

\subsubsection{Sensitivity to Rank $r$}

The rank parameter $r$ is a key hyperparameter in the Interference-suppressed Component Decoupling stage, controlling the dimensionality of the primary alignment core retained during decomposition. Specifically, $r$ determines the relative proportion between the principal alignment components that capture shared cross-modal structures and the domain-specific residual components that reflect dataset-dependent variations. Table~\ref{tab:rank_sensitivity} reports the performance under different rank settings on the clustered CC12M splits.

As $r$ increases from 8 to 64, performance steadily improves and reaches the best average score of 41.8 at $r=64$. However, further increasing the rank to 128 leads to a slight performance drop. This trend indicates that an excessively small rank limits the capacity of the shared alignment subspace and may discard useful cross-modal alignment information. In contrast, an overly large rank may introduce redundant or noisy components stemming from residual differences among projectors.

Overall, the results suggest that a moderate rank achieves the best trade-off between capturing shared alignment structures and suppressing residual discrepancies across projectors.

\begin{table}[t]
\centering
\caption{Sensitivity of rank $r$ on clustered CC12M splits. Avg. is the average score across all benchmarks.}
\label{tab:rank_sensitivity}
\renewcommand{\arraystretch}{1.1}
\setlength{\tabcolsep}{10pt}
\begin{tabular}{c|ccccc}
\toprule
Rank $r$ & 8 & 16 & 32 & 64 & 128 \\
\midrule
Avg. & 37.4 & 40.8 & 41.0 & \textbf{41.8} & 40.5 \\
\bottomrule
\end{tabular}
\vspace{-3ex}
\end{table}

\subsubsection{Layer-wise Disparity Analysis}

\begin{figure}[t]
\centering
\includegraphics[width=\linewidth]{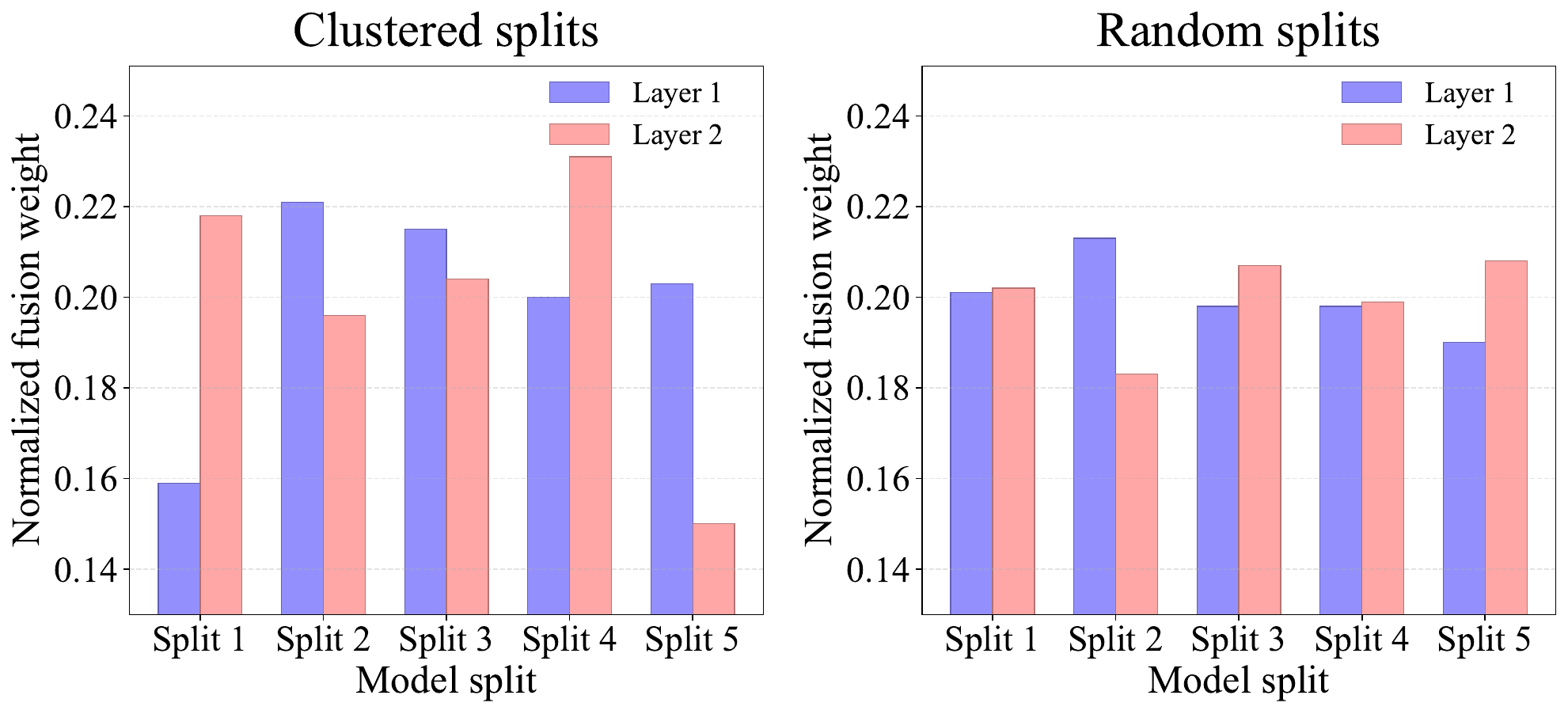}
\caption{
Layer-wise merge weights of expert models under the clustered and random settings.
Weights are obtained by converting cross-modal alignment scores into normalized merge coefficients using temperature-scaled softmax.
Different layers exhibit distinct weight distributions across experts, indicating clear layer-wise alignment contribution disparity.
}
\label{fig:layer_weight_difference}
\vspace{-2ex}
\end{figure}

To verify the existence of layer-wise alignment contribution disparity in projectors, we analyze the merge weights derived from the cross-modal alignment scores of different expert models.
Following Section~\ref{sec:method:layeraware}, the alignment scores are converted into normalized merge coefficients that determine each expert model's contribution during merging.
Figure~\ref{fig:layer_weight_difference} shows the normalized merge weights of the five expert models for each projection layer under both clustered and random settings.

Two observations can be made.
First, the dominant expert model varies across layers.
For example, under clustered splits, Split~2 receives the largest weight in Layer~1, while Split~4 becomes dominant in Layer~2.
Similar shifts appear in other splits, indicating that different projection layers favor different expert models and that alignment contributions are not uniformly distributed across layers.
Second, this disparity is more pronounced in the clustered setting.
When the pre-training data are partitioned by semantic clustering, the resulting projectors specialize in different alignment patterns, leading to larger variations in merge weights across layers.
In contrast, in the random setting, the weights are relatively balanced, reflecting smaller distribution discrepancies among datasets.

Overall, these results provide direct evidence of layer-wise alignment contribution disparity in projectors.
Since the ranking of expert models changes across layers, using a shared merge coefficient would ignore such differences and potentially degrade alignment quality.
This supports the necessity of the proposed Alignment-guided Layer-wise Merging, which assigns independent merge weights to each layer according to alignment strength.

\subsubsection{Consistency Analysis of CRF for Conflict Mitigation}

To better understand how Consistency-aware Residual Filtering (CRF) mitigates inter-model interference, we analyze the consistency of the domain-specific residuals across models before and after applying CRF.

As described in Section~\ref{sec:method:residual_filtering}, CRF operates only on the domain-specific residual branch $\mathbf{B}_i$, since cross-model conflicts mainly stem from directional inconsistencies in these components.
To quantify its effect, we measure the pairwise cosine similarity between the flattened domain-specific residuals of different models:
\begin{equation}
\text{Sim}(i,j) =
\frac{\langle \mathbf{B}_i, \mathbf{B}_j \rangle}{\|\mathbf{B}_i\| \|\mathbf{B}_j\|}.
\end{equation}

\begin{figure}[t]
\centering
\includegraphics[width=1.0\linewidth]{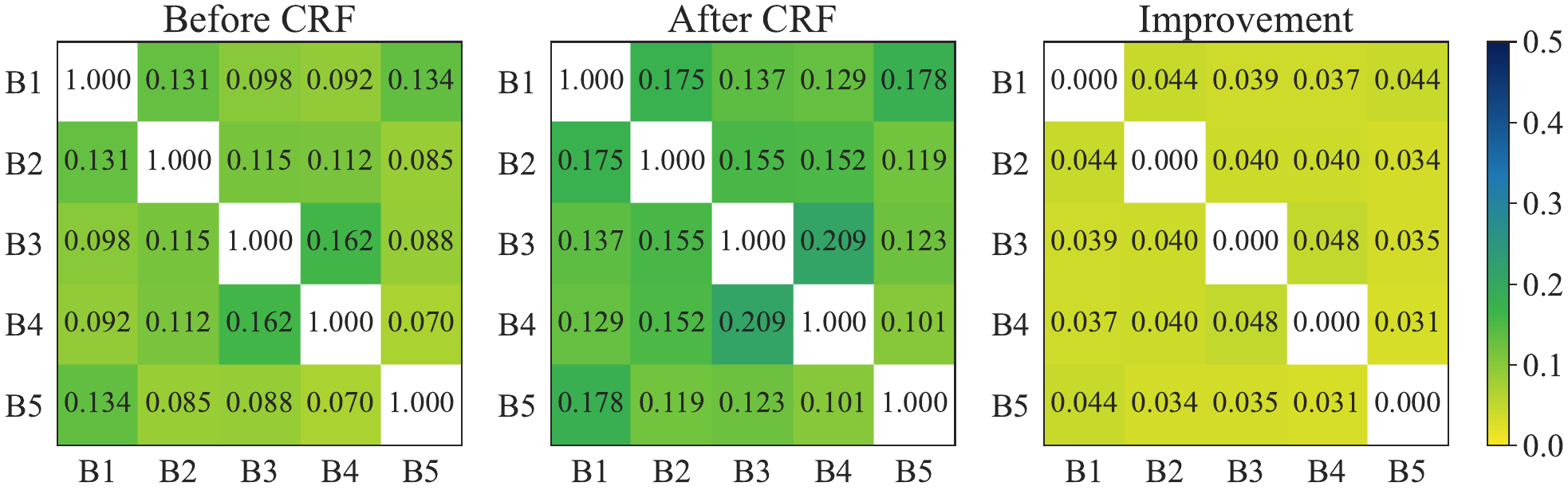} 
\caption{
Pairwise cosine similarity of the domain-specific residuals ($\mathbf{B}$) across five models under the clustered setting. 
Left: similarity before CRF. 
Middle: similarity after CRF. 
Right: improvement after CRF. 
CRF consistently increases inter-model consistency, indicating reduced conflicts in the residual subspace.
}
\vspace{-2ex}
\label{fig:crf_consistency}
\end{figure}

Higher similarity indicates that the residual transformations of two models are more geometrically aligned. 
Figure~\ref{fig:crf_consistency} shows the pairwise similarity matrices across the five models under the clustered setting. 
We observe a consistent increase in similarity across nearly all model pairs, indicating that CRF improves the consistency of the residual space.

These results show that CRF effectively suppresses inconsistent residual directions caused by distribution-specific variations. 
The increased consistency confirms that CRF reduces destructive interference in the domain-specific residual subspace, thereby reducing cross-projector conflicts and enabling more reliable merging.

\begin{table*}[!t]
\centering
\caption{Performance after LoRA fine-tuning with different merged projectors as initialization.
Best results are shown in bold.}
\label{tab:lora_finetune}
\resizebox{\textwidth}{!}{
\begin{tabular}{lccccccccc}
\toprule
Method 
& MMVet 
& MMBench$_{\text{EN}}$ 
& SEED 
& LLaVABench 
& POPE 
& MME-P 
& MMVP 
& MMStar 
& Avg. \\
\midrule
TIES + LoRA Fine-tuning
& \textbf{30.7} 
& 58.8 
& 60.1 
& 48.8 
& \textbf{87.4} 
& 1328.2 
& 53.0 
& 34.9 
& 55.0 \\

\rowcolor{oursblue}
\textbf{PivotMerge + LoRA Fine-tuning}
& 28.1 
& \textbf{59.6} 
& \textbf{61.5} 
& \textbf{53.9} 
& 86.3 
& \textbf{1336.4} 
& \textbf{61.0} 
& \textbf{35.5} 
& \textbf{56.6} \\
\bottomrule
\end{tabular}
}
\vspace{-3ex}
\end{table*}

\subsubsection{Principal Angle Analysis}
\label{sec:exp:principal_angle}

To examine whether PivotMerge reduces cross-projector interference, we analyze pairwise principal angles before and after applying PivotMerge.
For each projector, we build a model-level subspace from all projector layers: for the original models, it is constructed from the projector parameter updates $\Delta \mathbf{W}$, while after PivotMerge it is taken from the filtered coefficient matrix after Shared-space Decomposition.
For each projector pair, we compute the principal angles between the two subspaces and report their mean. Smaller principal angles indicate higher consistency across projectors and lower interference during merging.

\begin{figure}[t]
\centering
\includegraphics[width=0.89\linewidth]{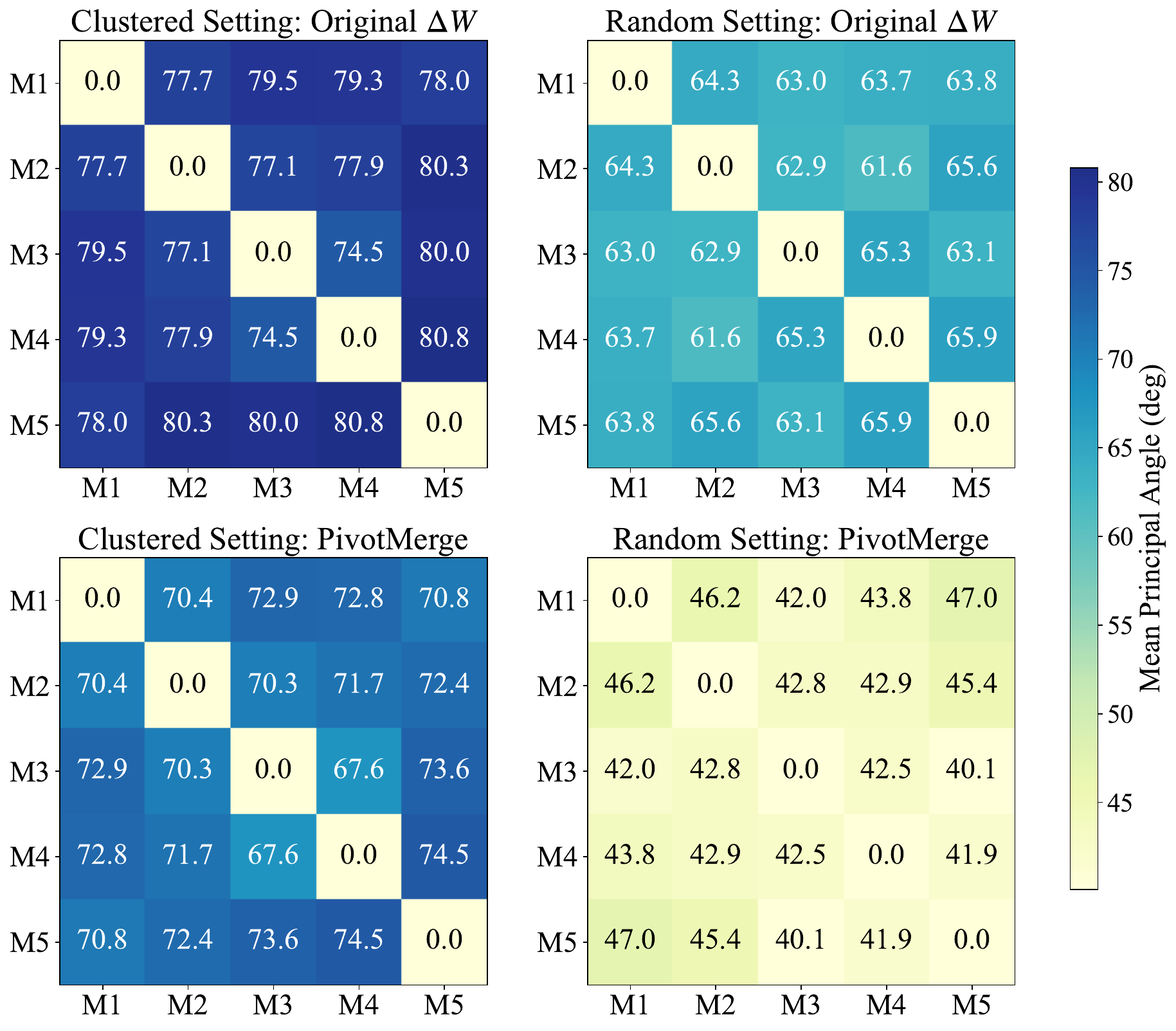}
\vspace{-1ex}
\caption{Pairwise mean principal angles before and after applying PivotMerge under clustered setting and the random setting.
The top row is computed from the original projector parameter updates $\Delta \mathbf{W}$, and the bottom row from the coefficient matrices after PivotMerge.
Smaller angles indicate higher consistency and lower interference.}
\label{fig:principal_angle_heatmaps}
\vspace{-3ex}
\end{figure}

Figure~\ref{fig:principal_angle_heatmaps} shows the pairwise mean principal angles under both clustered and random settings.
The effect differs across settings.
In the clustered setting, the original updates have much larger principal angles (around $78^\circ$ on average), reflecting stronger interference caused by heterogeneous pre-training data distributions.
After applying PivotMerge, the average angle decreases to about $71^\circ$, indicating reduced conflicts among highly heterogeneous projector updates.
In the random setting, the original updates start from a smaller average angle (around $64^\circ$), reflecting weaker heterogeneity.
PivotMerge further reduces the angles to about $43^\circ$, showing that consistency is further improved in this more homogeneous setting.
Because the initial interference is weaker, the downstream performance gain is correspondingly smaller.
Overall, PivotMerge consistently produces smaller principal angles than the original parameter space, demonstrating improved cross-projector consistency and reduced interference, especially in the clustered setting, where conflicts are more severe.

\subsubsection{Impact on Subsequent Fine-tuning}
\label{sec:exp:finetuning}

To further examine whether the quality of the merged projector affects subsequent fine-tuning, we conduct an additional experiment by following the LoRA fine-tuning stage of LLaVA-1.5 and initializing the models with different merged projectors.
Specifically, we compare models initialized with projectors produced by TIES and PivotMerge.
The results are summarized in Table~\ref{tab:lora_finetune}.

We observe that projector initialization has a clear impact on the final fine-tuning performance. Although the TIES-based initialization performs slightly better on two benchmarks, PivotMerge achieves better results on six out of eight benchmarks after fine-tuning and improves the overall average by 1.6 points.
These results indicate that the projector obtained during multimodal pre-training directly influences subsequent fine-tuning effectiveness.
By providing a stronger projector initialization, PivotMerge offers a better foundation for subsequent fine-tuning, allowing the model to better preserve and exploit the complementary cross-modal alignment capabilities integrated at the pre-training stage.

\section{Conclusion}

In this work, we move model merging beyond conventional post-finetuning settings by introducing the post-alignment merging task for MLLMs.
Under a training-free projector-merging setting with the vision encoder and LLM frozen, this task targets the cross-modal alignment stage of multimodal pre-training and raises two key challenges in this setting: cross-domain parameter interference among heterogeneous projector updates and layer-wise alignment contribution disparity across projection layers.
To address these challenges, we propose \textbf{PivotMerge}, a post-alignment merging framework that incorporates Shared-space Decomposition and Filtering to mitigate cross-domain parameter interference among heterogeneous projector updates, and Alignment-guided Layer-wise Merging to address layer-wise alignment contribution disparity through adaptive layer-specific weighting.
We hope this work inspires further exploration of model merging at the multimodal pre-training stage.

\vspace{-1ex}

\bibliographystyle{IEEEtran}
\bibliography{references}

\clearpage
\appendices

\section{Implementation Details}

\subsection{Models}

We follow the default pre-training setup of LLaVA-1.5\cite{liu2024improved} in all experiments.
Under both the clustered and random settings, the projector is trained for one epoch with a batch size of 256, a learning rate of $1\times10^{-3}$, cosine learning rate decay, a warmup ratio of 0.03, and a weight decay of 0.
We use the AdamW optimizer and DeepSpeed ZeRO-2 for distributed training.

\subsection{Benchmarks}

The dataset sizes and task focuses are summarized in Table~\ref{tab:benchmark_overview}, which mainly consists of several hundred to several thousand visual question answering (VQA), multiple-choice question (MCQ), and yes/no samples.

For fair comparison, all models are evaluated with \texttt{VLMEvalKit}\cite{duan2024vlmevalkit} under \texttt{bf16} precision.
Unless otherwise specified, all reported results follow the default evaluation settings of \texttt{VLMEvalKit}.
For overall average calculation, we normalize the MME-P results by dividing them by 20, following the setting in the main paper.

\begin{table}[H]
\centering
\caption{Overview of benchmarks used for evaluation.}
\label{tab:benchmark_overview}
\renewcommand{\arraystretch}{1.15}
\setlength{\tabcolsep}{2.5pt}
\footnotesize
\begin{tabular}{L{1.8cm} c c L{4cm}}
\toprule
\textbf{Benchmark} & \textbf{Category} & \textbf{Size} & \textbf{Capability Focus} \\
\midrule
MMVet~\cite{yu2023mm} & VQA & 218 & Integrated multimodal reasoning \\
MMBench$_{\text{EN}}$~\cite{liu2024mmbench} & MCQ & 6666 & General multimodal understanding \\
SEEDBench~\cite{li2023seed} & MCQ & 14232 & Broad perception and reasoning \\
LLaVABench~\cite{liu2023visual} & VQA & 60 & Visual instruction following \\
POPE~\cite{li2023evaluating} & Y/N & 5127 & Hallucination assessment \\
MME~\cite{fu2023mme} & Y/N & 2374 & Perception and cognition \\
MMVP~\cite{tong2024eyes} & MCQ & 300 & Fine-grained visual perception \\
MMStar~\cite{chen2024we} & MCQ & 1500 & Vision-indispensable reasoning \\
\bottomrule
\end{tabular}
\end{table}

\subsection{Merging Details}

In our implementation, bias terms are not merged as an independent branch.
Instead, for each linear layer, the bias vector is reshaped and appended as the last column of the corresponding weight matrix, forming an augmented matrix.
All subsequent operations, including \textit{Shared-space Decomposition and Filtering} and \textit{Alignment-guided Layer-wise Merging}, are performed jointly on this augmented matrix.
After merging, the last column is separated and restored as the bias term.
This design allows bias and weight parameters within the same layer to be processed in a shared subspace, reducing mismatch caused by separate handling.

For Consistency-aware Residual Filtering, the threshold $\tau$ in the sigmoid mask is not tuned independently in implementation.
Instead, we parameterize it by a retention ratio $\rho \in (0,1)$.
Given sorted consistency scores $c_{(1)} \le \cdots \le c_{(m)}$, we set
\[
k_{\tau} = \max\!\left(1,\left\lfloor m(1-\rho)\right\rfloor\right), \qquad
\tau = c_{(k_{\tau})},
\]
with boundary clipping when necessary.
Under the sigmoid gating function, larger consistency scores yield larger mask values, so $\rho$ controls the filtering strength through the induced threshold.
We use $\rho=0.5$ for clustered splits and $\rho=0.8$ for random splits.

For cross-modal alignment score weighting, we set the softmax temperature $\beta$ to 0.05, so that the differences among model weights remain within a reasonable range.

As the base merging method, we adopt TIES~\cite{yadav2023ties}.
For magnitude-based pruning methods such as TIES, we apply a special treatment: since model-scale information is not preserved in $\mathbf{V}_i^\top$ but remains in $\mathbf{\Sigma}$, we perform Component-wise Decoupling and the subsequent TIES-related operations on $\mathbf{\Sigma}\mathbf{V}_i^\top$.
During the masking stage, we do not apply additional hard top-$k$ sparsification after the sigmoid-based residual filtering, so all dimensions remain involved in the subsequent sign alignment and merging.
Furthermore, the scaling coefficients are normalized to a fixed sum of 5, corresponding to the number of expert projectors, to keep the overall parameter magnitude after merging unchanged.

\begin{figure*}[!t]
\centering
\includegraphics[width=0.93\textwidth]{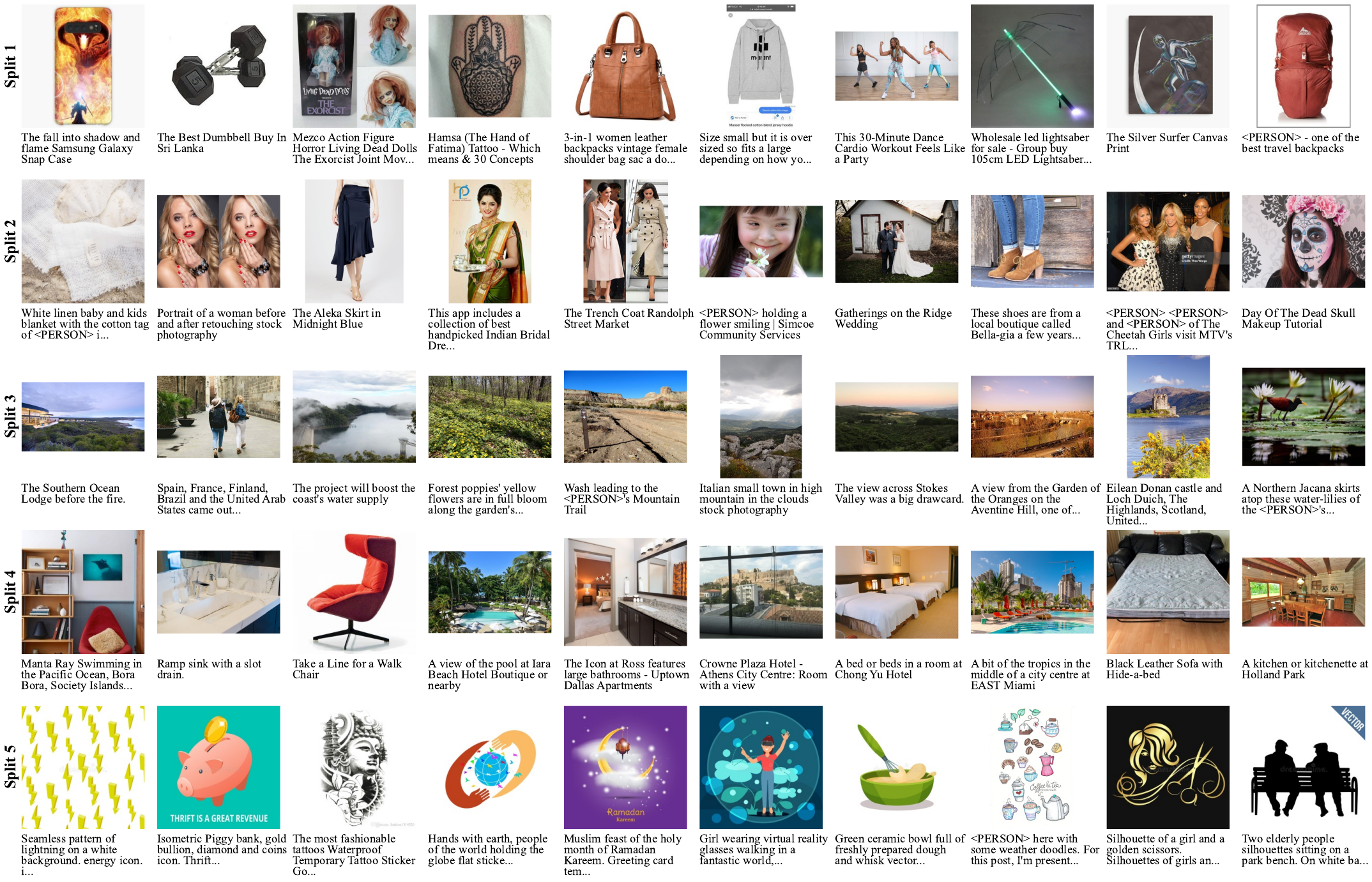}
\caption{Visualization of samples from clustered splits. Each row corresponds to one split. Samples within the same split exhibit strong semantic coherence, indicating clear domain-specific grouping.}
\label{fig:cluster_data_vis}

\vspace{2ex}

\includegraphics[width=0.93\textwidth]{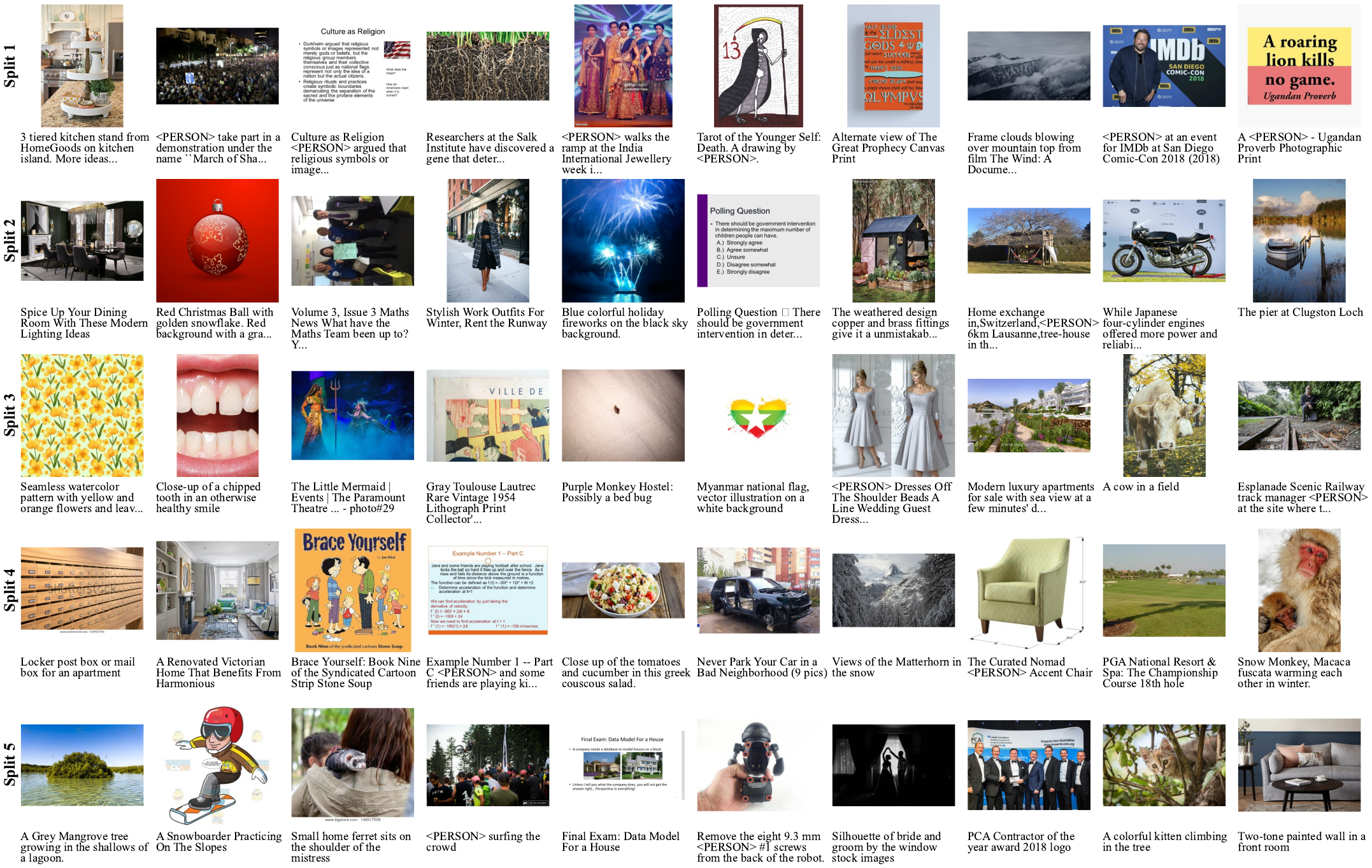}
\caption{Visualization of samples from random splits. Each row corresponds to one split. Samples are diverse and lack clear semantic grouping, indicating minimal domain discrepancy across splits.}
\label{fig:random_data_vis}
\end{figure*}

\subsection{Subsequent LoRA Fine-tuning Details}

For the subsequent fine-tuning experiment, we follow the default LoRA fine-tuning setup of LLaVA-1.5 and use its standard 665K instruction-tuning dataset.
For different initialization settings, we replace the projector with the merged projector produced by PivotMerge or TIES before LoRA fine-tuning.
LoRA is enabled with rank $128$ and $\alpha=256$.
The base learning rate is $2\times10^{-4}$, and the projector learning rate is $2\times10^{-5}$.
Training runs for one epoch with a per-device batch size of 16, cosine learning rate decay, a warmup ratio of 0.03, and a weight decay of 0.
We use \texttt{bf16} precision and DeepSpeed ZeRO-3 for distributed training.
Other settings follow the default LLaVA-1.5 configuration.

\section{Visualization of Data Splits}
\label{sec:exp:data_vis}

To better understand the distributional characteristics of different pre-training splits, we visualize randomly sampled image--text pairs from both the clustered and random splits, as shown in Fig.~\ref{fig:cluster_data_vis} and Fig.~\ref{fig:random_data_vis}.

As shown in Fig.~\ref{fig:cluster_data_vis}, the clustered splits exhibit clear semantic structures within each split.
Specifically, Split~1 is mainly dominated by daily objects and consumer products, Split~2 primarily contains human-centric content such as portraits, fashion, and social scenes, Split~3 is mainly composed of outdoor scenes including landscapes and natural environments, Split~4 focuses on indoor scenes such as rooms, furniture, and interior spaces, and Split~5 is dominated by icons, illustrations, and simple graphic drawings.
This strong intra-split consistency indicates that clustering introduces pronounced domain specialization, leading to substantial distributional heterogeneity across splits.

In contrast, Fig.~\ref{fig:random_data_vis} shows that the random splits do not exhibit such semantic grouping.
Each split contains a mixture of diverse content, including objects, people, scenes, and graphics, with no clear domain-specific patterns.
This results in relatively homogeneous data distributions across splits, with significantly reduced inter-split discrepancy.

These observations qualitatively support our experimental design: the clustered splits exhibit clearer semantic grouping and stronger heterogeneity, likely leading to more severe merging conflicts, whereas the random splits are more mixed and homogeneous, implying milder conflicts.
This contrast provides a suitable basis for evaluating the robustness of projector merging methods under different levels of heterogeneity.

\begin{figure*}[!t]
\centering
\includegraphics[width=\linewidth]{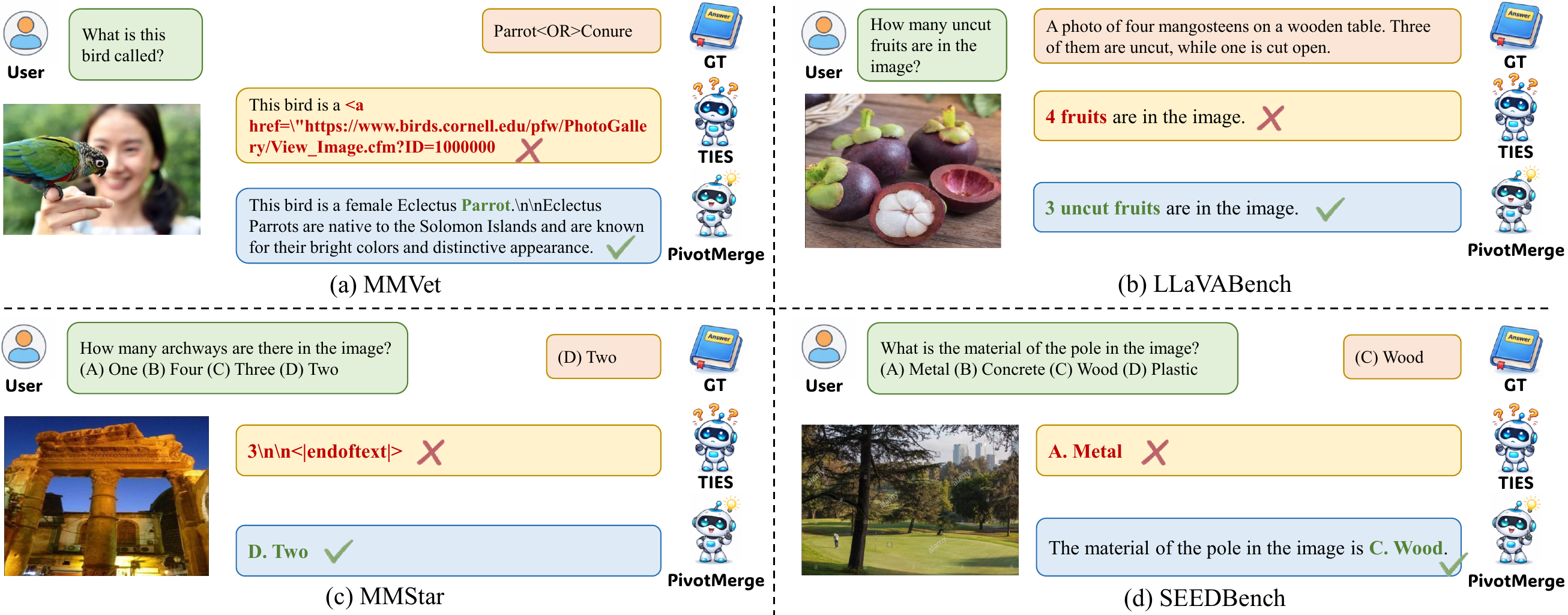}
\caption{Qualitative comparison between TIES and PivotMerge on multiple benchmarks. PivotMerge produces more accurate and grounded responses, while TIES exhibits errors such as hallucinated content, incorrect counting, and misclassification.}
\label{fig:case_studies}
\end{figure*}

\section{Case Studies}

To further examine the effectiveness of PivotMerge, we present qualitative case studies across multiple benchmarks in Fig.~\ref{fig:case_studies}.
We compare PivotMerge with TIES, a representative merging baseline, to analyze their behavior in different scenarios.

\textbf{(a) MMVet.}
In the visual recognition example, TIES produces a response containing irrelevant content, including an external hyperlink and other artifacts, indicating hallucination and poor grounding.
In contrast, PivotMerge correctly identifies the object as an Eclectus parrot and provides a concise and grounded description.
This suggests that PivotMerge better preserves the cross-modal correspondence between visual features and language representations.

\textbf{(b) LLaVABench.}
For the counting task, TIES incorrectly predicts the total number of fruits without distinguishing between cut and uncut instances, demonstrating limited fine-grained reasoning.
PivotMerge, however, correctly counts the number of uncut fruits, indicating more accurate alignment between visual details and language queries.

\textbf{(c) MMStar.}
In the multiple-choice reasoning example, TIES generates an incorrect answer with formatting artifacts, reflecting unstable decoding and unreliable reasoning.
PivotMerge correctly selects the answer ``(D) Two'', showing more reliable prediction in this case.

\textbf{(d) SEEDBench.}
For material recognition, TIES predicts an incorrect category (metal), while PivotMerge correctly identifies the pole as wood.
This demonstrates that PivotMerge can better capture subtle visual cues and maintain semantic consistency.

\textbf{Analysis.}
Across all examples, TIES tends to suffer from hallucination, counting errors, and incorrect semantic predictions, whereas PivotMerge consistently produces more accurate and grounded outputs.
This qualitative advantage is consistent with the design of PivotMerge, which incorporates Shared-space Decomposition and Filtering to mitigate cross-domain parameter interference among heterogeneous projector updates, and Alignment-guided Layer-wise Merging to address layer-wise alignment contribution disparity through adaptive layer-specific weighting.
As a result, PivotMerge better preserves common cross-modal alignment patterns while mitigating conflicts introduced by distribution-specific variations, leading to more stable and reliable predictions.

These qualitative results complement the quantitative results and further demonstrate the effectiveness of PivotMerge.

\end{document}